\documentclass{article}
\usepackage[final, nonatbib]{neurips_2022}
\usepackage[numbers]{natbib}
\usepackage[utf8]{inputenc} 
\usepackage[T1]{fontenc}    
\usepackage{hyperref}       
\usepackage{url}            
\usepackage{booktabs}       
\usepackage{amsfonts}       
\usepackage{nicefrac}       
\usepackage{microtype}      
\usepackage{wrapfig}
\usepackage{graphicx}
\usepackage{float}
\usepackage{hyperref}

\title{Scalable Neighborhood-Based Multi-Agent Actor-Critic}

\author{
    \texttt{Rasmus Jensen}\\ \texttt{rajensen@student.ethz.ch} \\
    \texttt{Tim Goppelsroeder}\\ \texttt{ tgoppelsroed@student.ethz.ch}
}

\begin{document}

\maketitle

\begin{abstract}
   We propose MADDPG-K, a scalable extension to Multi-Agent Deep Deterministic Policy Gradient (MADDPG) that addresses the computational limitations of centralized critic approaches. Centralized critics, which condition on the observations and actions of all agents, have demonstrated significant performance gains in cooperative and competitive multi-agent settings. However, their critic networks grow linearly in input size with the number of agents, making them increasingly expensive to train at scale. MADDPG-K mitigates this by restricting each agent's critic to the 
$k$ closest agents under a chosen metric which in our case is Euclidean distance. This ensures a constant-size critic input regardless of the total agent count. We analyze the complexity of this approach, showing that the quadratic cost it retains arises from cheap scalar distance computations rather than the expensive neural network matrix multiplications that bottleneck standard MADDPG. We validate our method empirically across cooperative and adversarial environments from the Multi-Particle Environment suite, demonstrating competitive or superior performance compared to MADDPG, faster convergence in cooperative settings, and better runtime scaling as the number of agents grows. Our code is available at \url{https://github.com/TimGop/MADDPG-K}.
\end{abstract}

\section{Introduction}
Multi-Agent Reinforcement Learning (MARL) has shown significant potential across fields including robotics, traffic control, and game-playing, with recent results such as AlphaStar \cite{vinyals2019grandmaster} demonstrating its capacity to handle complex multi-agent environments. A cornerstone of many successful MARL approaches is the centralized critic architecture, most prominently exemplified by Multi-Agent Deep Deterministic Policy Gradient (MADDPG) \cite{MADDPG}, in which each agent's critic is conditioned on the observations and actions of all agents in the environment. This design addresses the non-stationarity that arises when multiple agents learn simultaneously, and has been shown to substantially outperform independent learning baselines.
However, centralized critics carry a significant scalability cost, namely that the input dimensionality of each critic network grows linearly with the number of agents, making every forward and backward pass more expensive as the environment scales. This limits the practical applicability of MADDPG to environments with relatively few agents, and motivates the need for more efficient alternatives.
In this paper we propose MADDPG-K, which resolves this bottleneck by providing each agent's critic only with information from the
$k$ nearest agents under the Euclidean metric, rather than all agents in the environment. This reduces the critic input to a fixed size regardless of
$n$, while preserving the locality of information that is most relevant to each agent's decision. We note that the Euclidean metric is a natural choice for the spatial environments we consider, but the approach generalizes in a straightforward manner to any metric, or to a learned embedding function that maps agent state to a space in which proximity reflects behavioral relevance.

We also briefly report a negative result from an alternative approach we explored. For this approach we used a shared centralized critic updated across groups of cooperative agents using a sliding window scheme. This approach failed to achieve competitive performance in preliminary experiments and was not pursued further.
The remainder of the paper is structured as follows. Section 2 reviews related work. Section 3 describes the MADDPG-K algorithm and provides a complexity analysis. Section 4 presents experimental results across three MPE environments at varying agent counts, and Section 5 concludes with directions for future work.

\section{Related Work}
The amount of research in deep reinforcement learning increased significantly after the seminal DQN paper \cite{DQN} by researchers at DeepMind, which taught agents to play Atari games using Q-learning, a popular reinforcement learning algorithm combined with a deep neural network that estimates Q-values. \\
Research in the years following established methods to extend DQNs to be able to handle continuous action spaces using an actor-critic approach. This led to the development of algorithms such as Deep Deterministic Policy Gradient (DDPG)\cite{DDPG}, Twin Delayed Deep Deterministic Policy Gradient (TD3)\cite{TD3}, and Soft Actor Critic (SAC)\cite{SAC}.\\\\
A naive extension of these algorithms to MARL environments is possible in the form of independent Q-learning \cite{Tan1997}, however recent research such as the MADDPG paper \cite{MADDPG} outlines a more informed approach, where in MADDPG the DDPG algorithm is extended by augmenting each agent's critic input with the critic input that all other agents would traditionally receive (i.e., all agents' observations and actions are fed into each agent's critic). This is done to alleviate the non-stationarity of the Markov Decision Process when multiple agents are involved. MADDPG showed improved performance on all environments tested in the paper \cite{MADDPG} over DDPG, especially on their cooperative communication and predator-prey environments (i.e. tag).
Recent advancements in the field have integrated MADDPG and SAC, leveraging the attention mechanism to facilitate the joint training of critics. This innovative approach is exemplified by the MAAC (Multi-agent Actor-Attention-Critic) algorithm \cite{attention-critic}, which samples actions from all target actors, as opposed to just the current agent. This modification has demonstrated enhanced performance across various test environments. Other non-attention based variants of this approach exist and have themselves shown to be quite capable \cite{rashid2020monotonic, peng2021facmac}.\\\\
A major drawback of most algorithms that use a centralized critic approach like MADDPG is that the input to the agents' critics scales linearly with the number of agents present in the environment. This is because the critic receives the observations and actions of all agents in the environment.
In our project, we aim to remedy this by restricting the critic's input for a given agent to the $K$ agents closest to that agent, according to the Euclidean distance between them. This helps to focus on more relevant information if the correct metric is used, and also reduces complexity, allowing the algorithm to scale to environments with larger numbers of agents.\\
An alternative approach to this is given by \cite{meanField}, where the critic's scaling problem is solved by averaging over pairwise local interactions between a given agent and other agents in a neighborhood around the given agent.\\\\
Concurrently to the work on the MADDPG, a separate group developed a similar approach\cite{CMPG}, also with a centralized critic, but using a counterfactual baseline. This baseline estimates the expected return if an agent were to take a different action while other agents maintain their current actions. By comparing an agent's actual return with this counterfactual baseline, COMA can potentially 
more accurately attribute credit for rewards. However high variance of the counterfactual baseline can cause the gradients for
updates to become unstable\cite{papoudakis2020comparative}. 
\\\\
Since we are working on an extension to MADDPG, we need access to the environments used in the MADDPG paper\cite{MADDPG}. However, the original environments are no longer maintained and contain obsolete code. Instead, we access them through Petting Zoo \cite{pettingzoo}, which is the multi-agent equivalent of openAI gym\cite{openAIgym} and contains a currently maintained version of these environments.

\section{Methods/Algorithms}
The use of a centralized critic, which obtains actions and observations from all agents as input can improve the efficiency of a MARL algorithm by dealing with the non-stationarity of the Markov decision process when multiple agents are involved, but adds additional complexity by making the input to the critic scale linearly with the number of agents in the environment.\\ Our first method reduces complexity, by when updating an agent, only focusing on a $K$ sized subset of agents that are closest to the agent in question with respect to the Euclidean distance in the environment. One could however also substitute the Euclidean metric for any other metric or even learn an embedding function and measure relatedness in embedding space using a chosen metric.\\ By focusing only on a subset of $K$ agents, we reduce the linear scaling of the critic input to a constant number of inputs, where $K$ is a hyperparameter. In addition, focusing only on the observations and actions of the closest agents can lead to faster learning in the environment by focusing the agent on more relevant information.
\subsection{Algorithm}
\includegraphics[width=10cm]{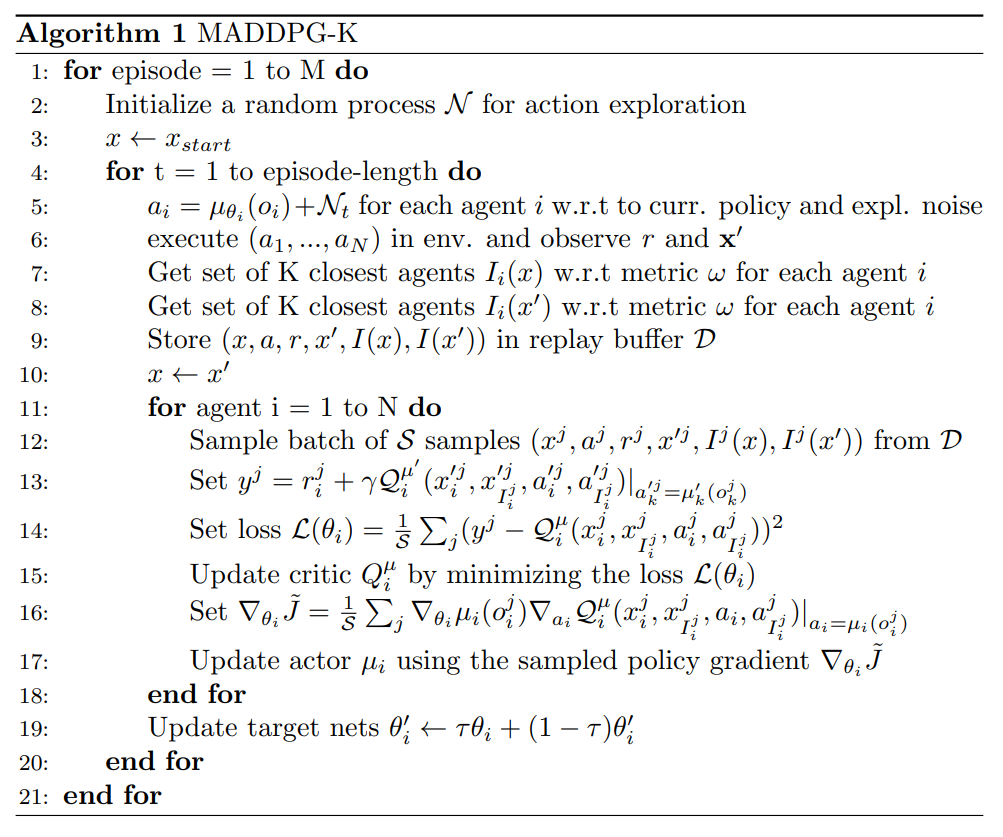}
\\
The main difference between MADDPG and MADDPG-K lies in the additional index sets for the current and next state that are additionally stored for each transition (i.e. $I(x)$ and $I(x')$). These index sets allow us to precompute the neighbors of given agents, which can then be used when updating these agents by feeding their critics only the observations and actions of their neighbors, where the agents' observations and actions are sorted according to their proximity to the given agent in the input vector.
\subsection{Complexity Analysis}
For the following section, we define the following variables:
\begin{itemize}
    \item $n$: number of agents
    \item $h$: hidden size
    \item $b$: batch size
    \item $l$: number of layers
    \item $a$: agent action space
    \item $o$: agent observation space
    \item $k \le n$: $k$-nearest neighbor setting
    \item $u$: update frequency
    \item $d$: dimension of the vector space in which the agents move (in our case, $d=2$)
\end{itemize}
We analyze the two algorithms by separating costs that arise at each environment step from those that arise at each update, since these occur at different rates and conflating them produces a misleading total expression.

\subsubsection{Per environment step}
Standard MADDPG incurs no additional cost per environment step beyond environment interaction itself.
MADDPG-K must additionally compute the $k$-nearest neighbors for each agent. Naively, this requires computing all $n(n-1)/2$ pairwise Euclidean distances in $d$-dimensional space, giving a cost of $\mathcal{O}(dn^2)$ per step. The resulting index sets $I(x)$ and $I(x’)$, which record the $k$ closest agents to each agent at the current and next state, are stored directly in the replay buffer alongside the transition. This precomputation adds $\mathcal{O}(nk)$ memory per stored transition.

\subsubsection{Per Update Step}
During each update, both algorithms process a batch of $S$ transitions sampled from the replay buffer.
The decentralized actor networks are identical in both algorithms. Each processes a single agent's observation to produce an action, giving a total actor cost across all agents of $\mathcal{O}(nb(oh + lh^2 + ha))$.
The critical difference lies in the centralized critic. In standard MADDPG, each agent's critic receives the observations and actions of all $n$ agents, so the critic input has dimension $n(o + a)$. The total critic cost for the batch is therefore $\mathcal{O}(nb(n(o + a)h + lh^2))$, with the $n^2$ factor arising from the matrix multiplication at the first critic layer.
In MADDPG-K, because neighbor indices were precomputed and stored at collection time, retrieving the relevant observations and actions at update time requires only $\mathcal{O}(nkb)$ index lookups. The distance computation cost is thus charged entirely to the per-step budget above.
The critic input is then of dimension $k(o + a)$, giving a total critic cost of $\mathcal{O}(nb(k(o + a)h + lh^2))$. The $n$ dependence in the neural network operations has been reduced to $k$, which is a fixed constant independent of the total agent count.
\subsubsection{Total Training Complexity}
Combining the per-step and per-update costs over a training run of $T$ total environment steps, with updates occurring every $u$ steps:
\begin{itemize}
    \item MADDPG: $\mathcal{O}((T/u) \cdot nb(lh^2 + h(n(o + a) + o + a)))$
    \item MADDPG-K: $\mathcal{O}(T \cdot dn^2 + (T/u) \cdot nb(lh^2 + h(k(o + a) + o + a)))$
\end{itemize}
Written this way, the structural difference becomes clear. MADDPG's dominant cost is a neural network operation that scales in terms of $n^2$. It appears inside the critic forward and backward passes, which are executed every update. MADDPG-K's $n^2$ cost is the per-step distance computation, which involves only scalar arithmetic and is bounded by a small constant ($d=2$ in our experiments).

\subsubsection{Practical Implications}
Both algorithms share $\mathcal{O}(n^2)$ asymptotic complexity with respect to the number of agents, but this similarity at the level of Big-O notation obscures a practically important distinction: the operations driving the quadratic term are fundamentally different in cost.
In MADDPG, the $n^2$ factor enters through large matrix multiplications inside the critic network, executed once per update step across a batch of size $b$. As $n$ grows, every forward and backward pass through the critic becomes more expensive in proportion to the compute intensity of deep network operations.
In MADDPG-K, the $n^2$ factor enters only through pairwise distance computation which involves scalar subtraction and addition in two dimensions and is performed once per environment step. These operations are orders of magnitude cheaper per unit than neural network matrix multiplications, and are easily parallelized. Once $k$ is fixed, the critic's computational cost is constant regardless of how large $n$ becomes.
This distinction is confirmed empirically in \ref{fig:scalability}, which shows measured wall-clock time per 100 episodes as a function of $n$ on identical hardware and network architectures. MADDPG-K requires less time per episode as $n$ grows, with the gap widening beyond $n = 45$. We note additionally that, because MADDPG-K's critic operates on a fixed-size neighborhood, it does not require as wide a network as MADDPG to achieve comparable performance, a secondary benefit not reflected in the runtime comparison of \ref{fig:scalability}, where both algorithms use identical architectures.

\section{Experiments/Results/Discussion}
\subsection{Environments}
We compared our algorithm to MADDPG on a variety of multi-agent environments from the MPE environment set \cite{mordatch2018emergence}. To test how MADDPG-K performs when observing only a diminishing fraction of agents, we increased the number of agents in some of the environments that allowed for this adjustment. The main environments used in this project are illustrated and described below, and all are scalable to larger numbers of agents, which is not generally the case for all MPE environments.
\begin{figure}[H]
\minipage[t]{0.32\textwidth}
  \includegraphics[width=\linewidth]{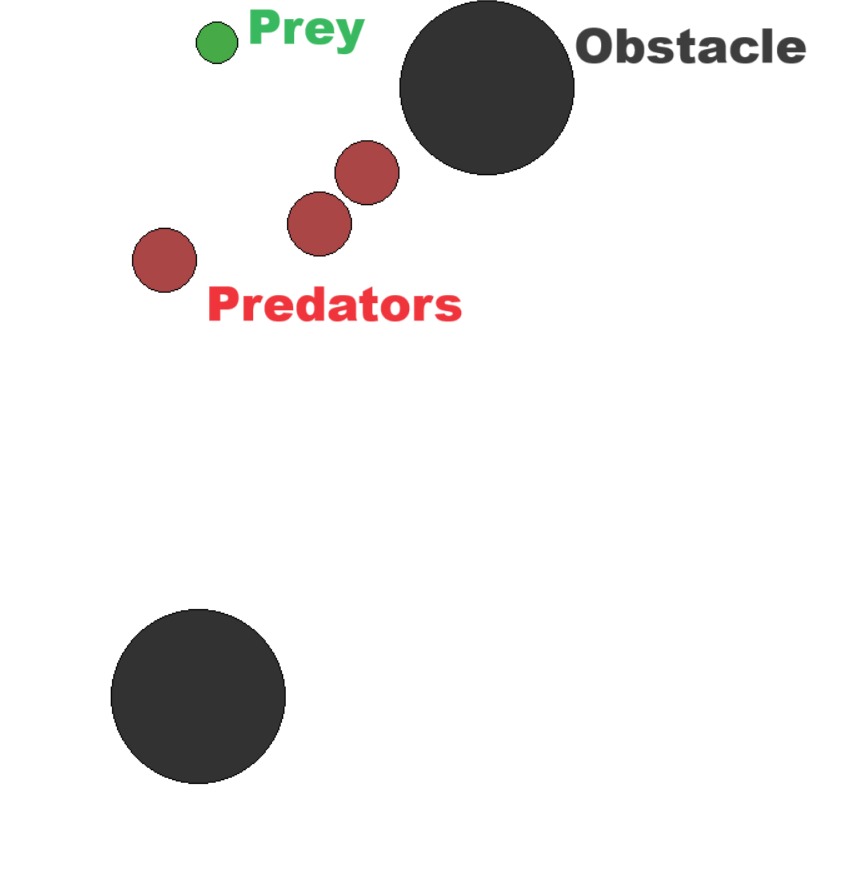}
  \caption{Simple Tag}\label{fig:simple_tag}
\endminipage\hfill
\minipage[t]{0.32\textwidth}
  \includegraphics[width=\linewidth]{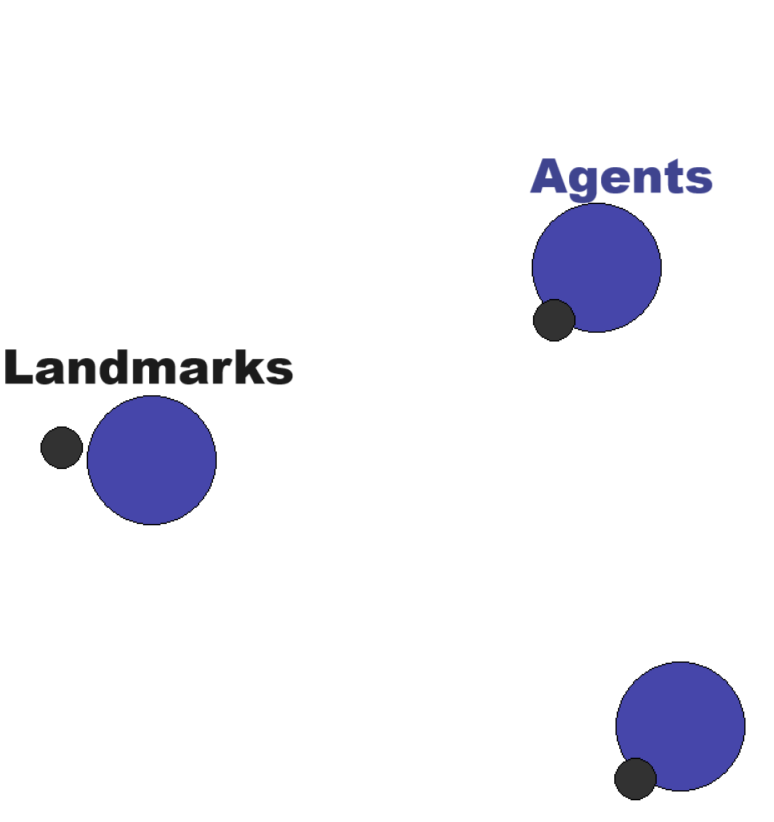}
  \caption{Simple Spread}\label{fig:simple_spread}
\endminipage\hfill
\minipage[t]{0.32\textwidth}
  \includegraphics[width=\linewidth]{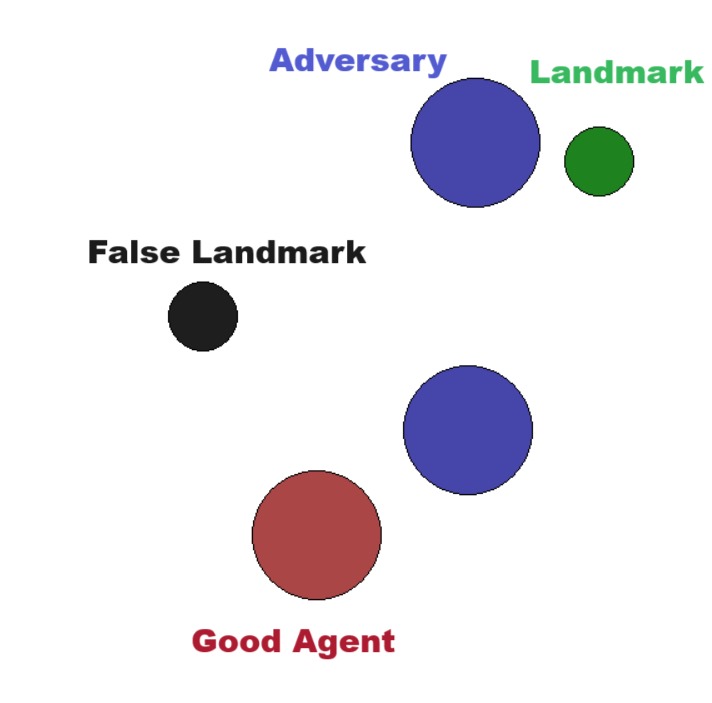}
  \caption{Simple Adversary}\label{fig:simple_world_comm}
\endminipage
\end{figure}
\subsubsection*{Simple Tag}
In a simulated predator-prey environment, faster green "Good" agents are pursued by slower red "Adversary" agents within a square arena. Each time an Adversary tags a Good agent, the Good agent incurs a -10 penalty, while the Adversaries gain a +10 reward(globally rewarded).  Obstacles in the arena hinder both types of agents.  To prevent Good agents from simply fleeing the arena, an exponential penalty is imposed based on their distance outside the boundaries.
\subsubsection*{Simple-Spread}
This environment features N agents and N landmarks (default N=3). The agents' main objective is to efficiently cover all landmarks while avoiding collisions. Agents receive a global reward based on the proximity of the closest agent to each landmark (i.e. sum of minimum distances). To encourage safer navigation, agents will incur a local penalty of -1 for each collision with another agent.
\subsubsection*{Simple Adversary}
In this environment there is a single adversary (red), N good agents (green), and N landmarks (default N=2). All agents observe the position of the landmarks and the other agents. One landmark is designated as the ‘target landmark’ (colored green). Good agents are rewarded based on the proximity of the closest good agent(globally rewarded) to the target landmark, but are negatively rewarded based on how close the adversary is to the target landmark. The adversary is rewarded based on its distance to the target, but it doesn’t know which landmark is the target landmark. All rewards are given by the unscaled Euclidean distance. This means that good agents must learn to "split up" and cover all landmarks to deceive the adversary.
\subsection{Results:}
\begin{figure}
    \centering
    \includegraphics[width=\textwidth]{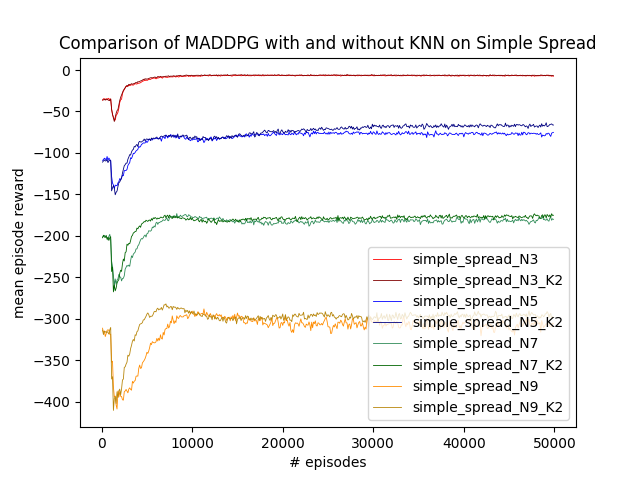}
    \caption{MADDPG-K and MADDPG on Simple Spread}
    \label{fig:maddpg-k_spread_comp}
\end{figure}

Before extending our  implementation to MADDPG-K we tested our implementations of independent DDPG and MADDPG on the simple tag environment, as well as the DDPG and MADDPG implementations from the paper (by using the code on their GitHub). We evaluated each algorithm by running it 10 times and calculating the mean learning trajectory (as depicted in \ref{fig:simple_tag}). Finally, the sum of rewards is the sum across all four agents in the environment.
Both implementations converge to higher rewards than the implementations from the original MADDPG paper.
We believe that the improved performance is primarily due to the initial random action exploration (maximum entropy principle), which leads to more consistent learning. Certain other changes, such as the size of our replay memory being $10^5$ rather than $10^6$, allow agents to focus on more recently stored transitions when updating, and may also have influenced the performance in this environment.
\begin{figure}[H]
\minipage{\textwidth}
  \includegraphics[width=\linewidth]{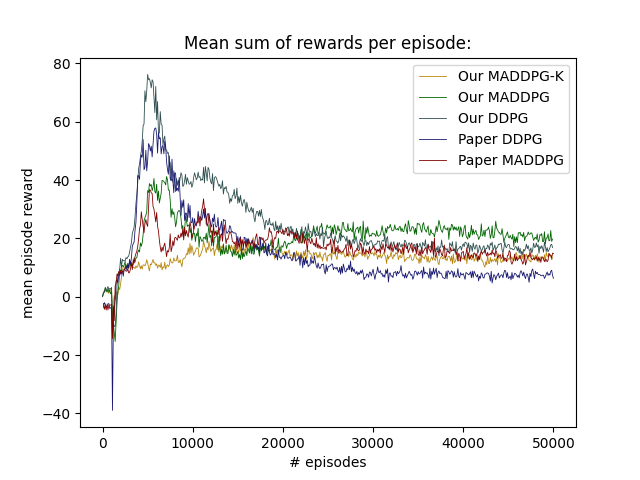}
  \caption{Simple Tag}\label{fig:simple_tag}
\endminipage
\end{figure}
We also ran our MADDPG-K implementation on the Simple Tag environment. 
This is also depicted in \ref{fig:simple_tag} and shows that while it does not converge to a higher reward than our MADDPG implementation, it is competitive with the MADDPG implementation from the paper.
One can also see that the learning is smoother for MADDPG-K. This might be due to the ordering of adversaries or agents according to proximity. The ordering simplifies the goal of the agents, since they do not have to find out which agent is closest to them in their observation space. Instead, they can assume that the first observation in their observation space is always closest and adjust their actions accordingly.
No significant difference was found between our MADDPG and MADDPG-K on Simple Tag when displaying the environment.\\\\
Because the results are less clear in adversarial environments (i.e., higher total rewards don't necessarily indicate better performing agents), we decided to see how the performance of MADDPG-K would evolve as the number of agents was increased in a cooperative environment. To do so, we chose the Simple Spread environment and scaled the number of agents from $3$ to $9$ while keeping $K=2$ (i.e. we look at the 2 closest neighbors as well as the current agent) and plotted the results \ref{fig:maddpg-k_spread_comp}.
From the results we can clearly see that MADDPG-K achieves equal or better rewards for all different sizes of $N$ tested. For $N=3$ the performance is near identical. This is expected, because the MADDPG-K agent is not constrained to a subset of agents. However, once $K$ is smaller than $N$, MADDPG-K seems to learn faster and converge to larger rewards than MADDPG. We hypothesize that this is due to the fact that agents in MADDPG-K benefit from focusing on agents that are closer to them, since the other agents are less important to that agent's goal.
This advantage diminishes as the number of agents increases. The largest advantage of MADDPG-K over MADDPG on Simple Spread was for $N=5$ and $K=2$, after which the advantage continues to decrease and at a certain point we hypothesize that the loss of information will become prohibitive, allowing standard MADDPG to achieve superior performance. This indicates that the choice of the hyperparameter $K$ has a large effect on the performance of the algorithm and the optimal choice of $K$ is environment dependent.
\begin{figure}
    \centering
    \includegraphics[width=\textwidth]{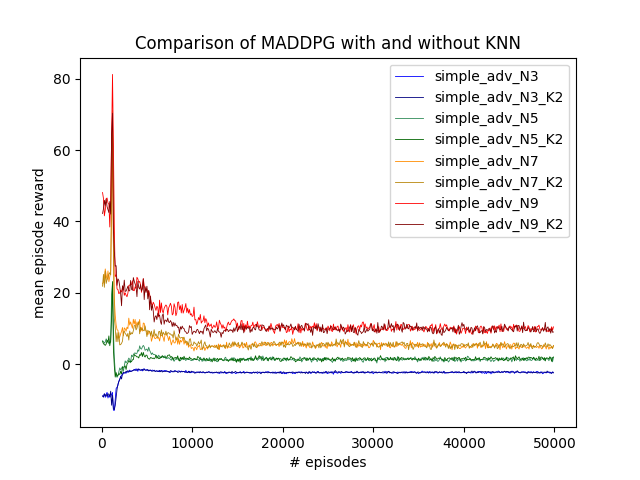}
    \caption{MADDPG and MADDPG-K on Simple Adversary}
    \label{fig:maddpg-k_simp_adv_comp}
\end{figure}
\newpage
For the environment Simple Adversary we conducted the same experiment \ref{fig:maddpg-k_simp_adv_comp} comparing MADDPG and MADDPG-K as we did on Simple Spread, where we fix $K=2$ and scale $N$ from $3$ to $9$.
MADDPG-K demonstrates competitive performance when compared to MADDPG, even as $N$ increases. No significant differences were observed between MADDPG and MADDPG-K. One minor difference seems to be that MADDPG-K converges faster on average than MADDPG. However more runs would have to be conducted to evaluate this statement properly. One should also note that, in the Simple Adversary Environment, only the number of Agents/Landmarks can be scaled as the number of Adversaries is fixed at $1$. 
Since agents can receive positive reward while adversaries can only minimize their negative reward this results in higher total average rewards when the number of agents in the environment increases.\\\\
In summary MADDPG-K shows improvements over MADDPG on cooperative environments such as converging to a higher reward and learning faster. On adversarial environments MADDPG-K still seems to be quite competitive with MADDPG in terms of reward and like in the cooperative environments MADDPG-K tends to learn faster than MADDPG.

\newpage
\section{Conclusion/Future Work}
This work presents MADDPG-K, an extension of the Multi-Agent Deep Deterministic Policy Gradient (MADDPG) algorithm \cite{MADDPG}, designed to mitigate the scalability limitations of MADDPG while also helping agents focus on more relevant information, which can improve an agent's ability to learn its policy. By utilizing a neighborhood-based network, MADDPG-K reduces computational complexity, allowing for the training of more complex multi-agent systems with larger agent populations. Results show that MADDPG-K is competitive with MADDPG on all the MPE environments tested, and in some specific environments, MADDPG-K even seems to outperform MADDPG.\
Looking ahead, MADDPG-K offers a promising solution for complex environments where traditional MADDPG might necessitate a more intricate network architecture to achieve comparable performance. We theorize, since MADDPG-K only focuses on the "more important" agents and thus seems to simplify the problem being learned, that it can achieve similar performance with a less complex network architecture than MADDPG for the same environment.\
Furthermore, while MADDPG-K significantly reduces the complexity of the neural network updates, the current implementation relies on a brute-force pairwise distance calculation to determine the $k$-nearest neighbors, which retains an $\mathcal{O}(n^2)$ computational overhead. A highly promising avenue for future work is the integration of more efficient spatial partitioning data structures, such as KD-Trees or Spatial Hashing, which can reduce the neighborhood search complexity to $\mathcal{O}(n \log n)$ or even $\mathcal{O}(n)$ in low-dimensional environments. For higher-dimensional state spaces, Approximate Nearest Neighbor (ANN) algorithms could be utilized. Implementing MADDPG-K with these advanced search methods would entirely eliminate the remaining quadratic bottlenecks, potentially unlocking even more substantial speedups and allowing the algorithm to scale to vastly larger agent populations.\
Another direction for future work would be to use different metrics, each of which would induce a different neighborhood. A particularly interesting idea would be to introduce a learned embedding function that maps agent information to an embedding space where an appropriate metric, such as the Euclidean distance metric, can be used to establish relatedness. The use of such an embedding function could potentially lead to an improved measure of relatedness between agents, leading to better results, while also allowing the algorithm to be used in environments where the position of the agent in space is not particularly relevant or does not exist.\
We leave these investigations to future work.

\newpage
\medskip
\bibliographystyle{plainnat}
\bibliography{ref.bib}


\clearpage
\appendix
\section{Appendix}
\subsection{Configurations}
For both our DDPG and MADDPG implementations, we use the following hyperparameters and settings:
\begin{itemize}
    \item Adam optimizer with a learning rate of 0.01
    \item $\tau$=0.01 (target network update parameter)
    \item $\gamma=0.95$
    \item replay memory size is $10^5$
    \item update after every 100 samples collected per agent
    \item batch size of 1024
\end{itemize} 
These hyperparameters have been chosen to match those in the MADDPG paper\cite{MADDPG} to allow for a better comparison with our algorithm. In addition, unlike the paper, we sample random actions uniformly before the first update of the networks and have a replay memory that is smaller by one factor of $10$. We do this to better explore the environment before the first update using the principle of maximum entropy and have a smaller replay memory to not keep transitions that are too old(i.e. from a time when the policy was not particularly "good" yet).
\subsection{Additional Plots}
Below we include further plots showing the individual performance of agents and adversaries on the base version of simple tag environment for each algorithm.
\begin{figure}[H]
    \centering
    \includegraphics[width=6cm]{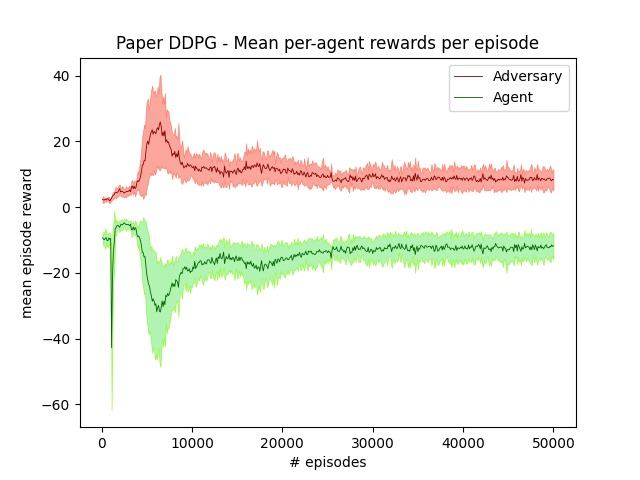}
    \includegraphics[width=6cm]{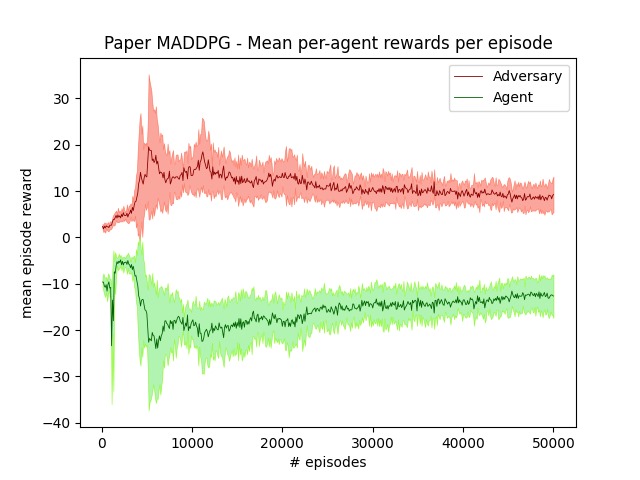}
    \caption{Algorithms from MADDPG paper GitHub}
    \label{fig:paper}
\end{figure}
\begin{figure}[H]
    \centering
    \includegraphics[width=6cm]{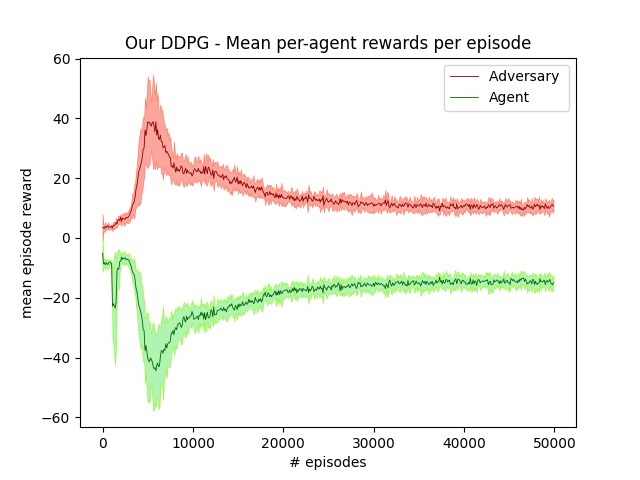}
    \includegraphics[width=6cm]{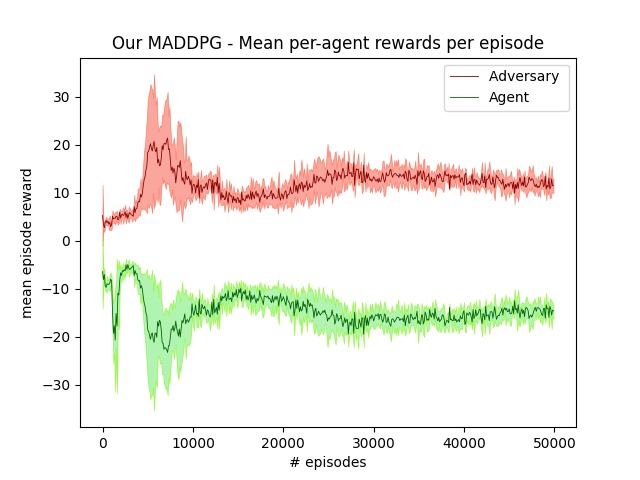}
    \caption{our versions of DDPG and MADDPG}
    \label{fig:paper}
\end{figure}
\begin{figure}[H]
    \centering
    \includegraphics[width=6cm]{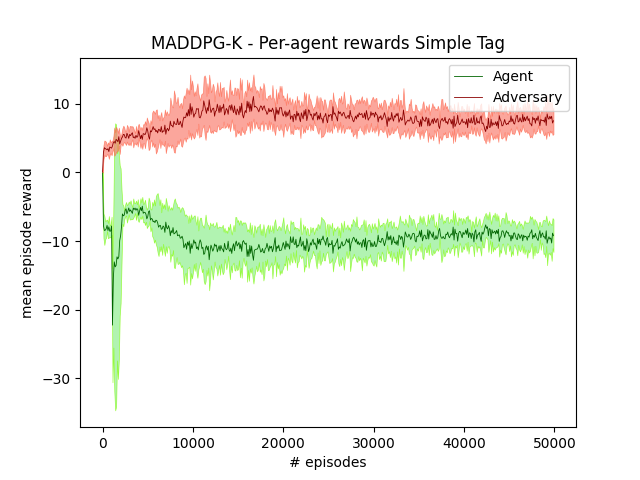}
    \caption{MADDPG-K 1 agent 3 adversaries K=3}
    \label{fig:maddpg-k_tag_ag_adv}
\end{figure}

\begin{figure}[H]
    \centering
    \includegraphics[width=6cm]{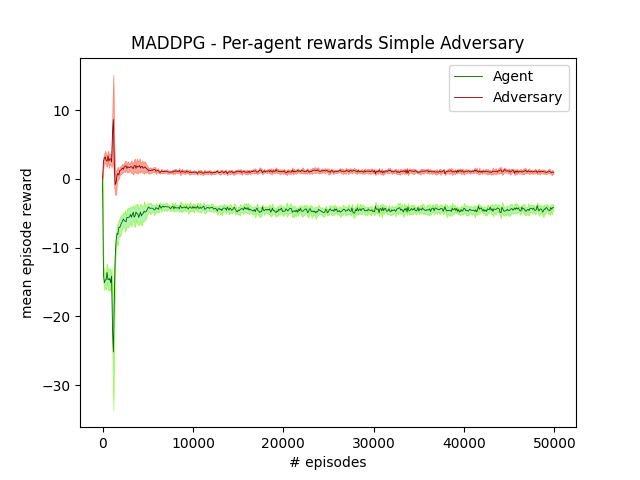}
    \includegraphics[width=6cm]{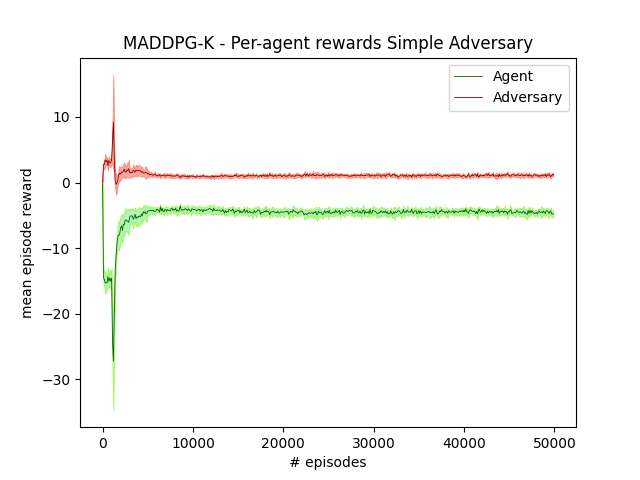}
    \caption{MADDPG and MADDPG-K Simple Adversary N=3, K=2}
    \label{fig:maddpg_maddpg_k_simp_adv_n3}
\end{figure}
\begin{figure}[H]
    \centering
    \includegraphics[width=6cm]{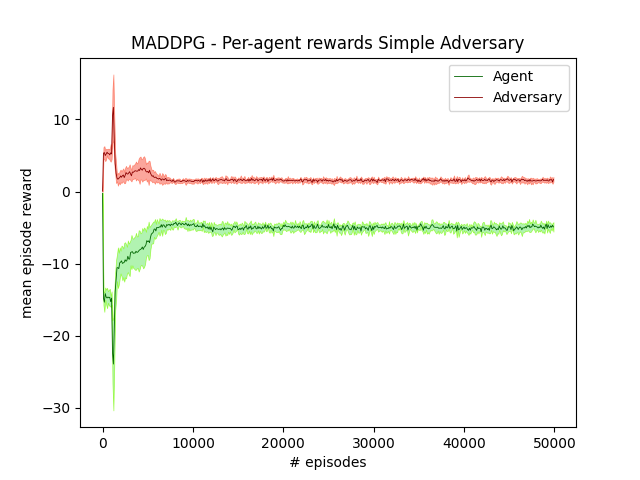}
    \includegraphics[width=6cm]{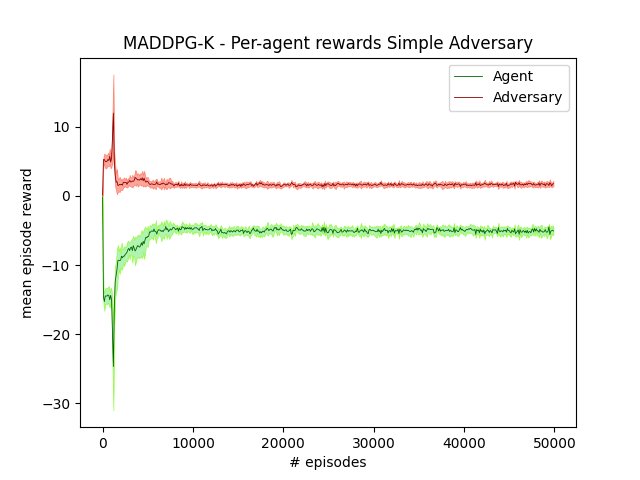}
    \caption{MADDPG and MADDPG-K Simple Adversary N=5, K=2}
    \label{fig:maddpg_maddpg_k_simp_adv_n5}
\end{figure}
\begin{figure}[H]
    \centering
    \includegraphics[width=6cm]{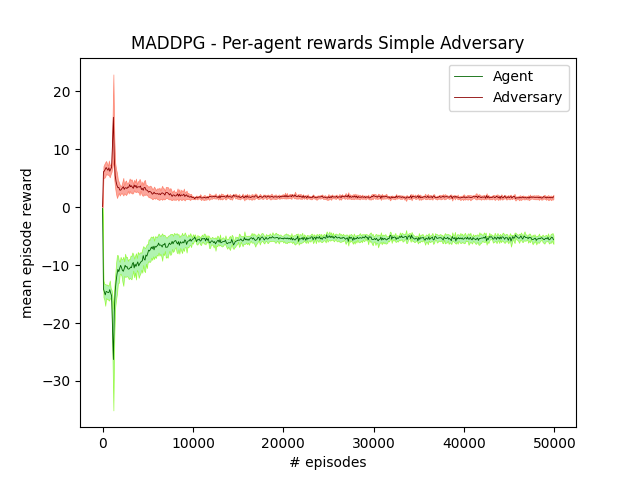}
    \includegraphics[width=6cm]{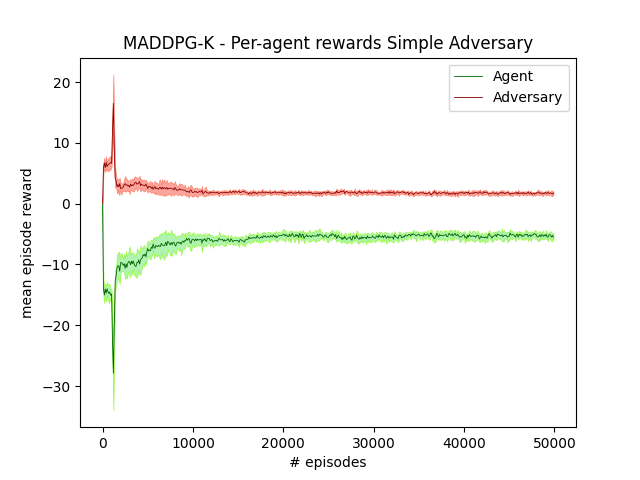}
    \caption{MADDPG and MADDPG-K Simple Adversary N=7, K=2}
    \label{fig:maddpg_maddpg_k_simp_adv_n7}
\end{figure}
\begin{figure}[H]
    \centering
    \includegraphics[width=6cm]{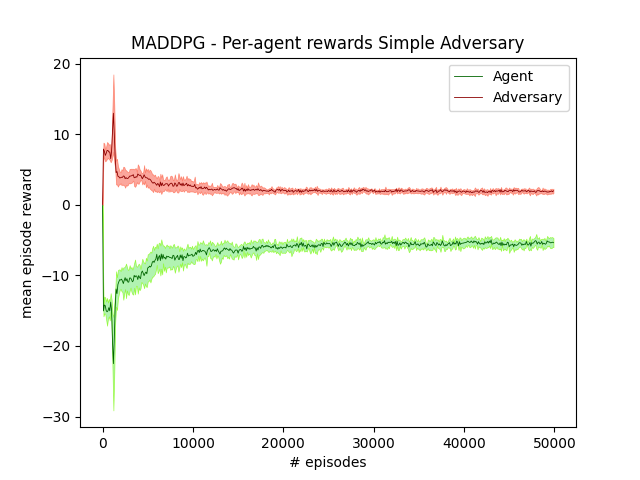}
    \includegraphics[width=6cm]{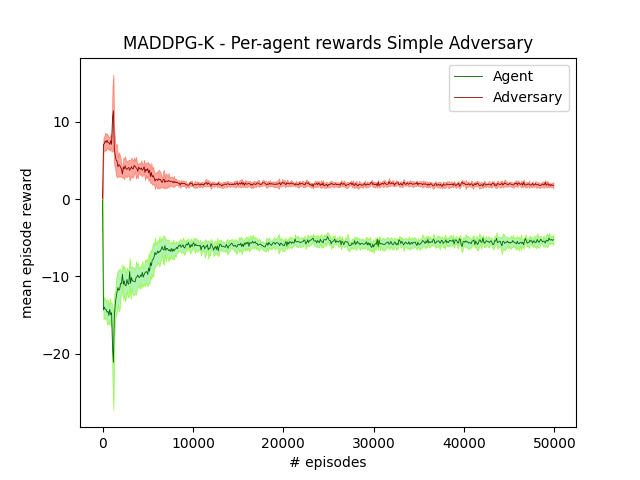}
    \caption{MADDPG and MADDPG-K Simple Adversary N=9, K=2}
    \label{fig:maddpg_maddpg_k_simp_adv_n9}
\end{figure}

\begin{figure}[H]
    \centering
    \includegraphics[width=6cm]{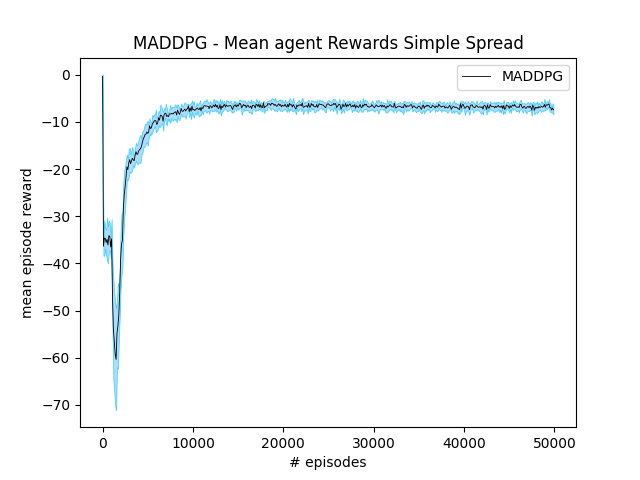}
    \includegraphics[width=6cm]{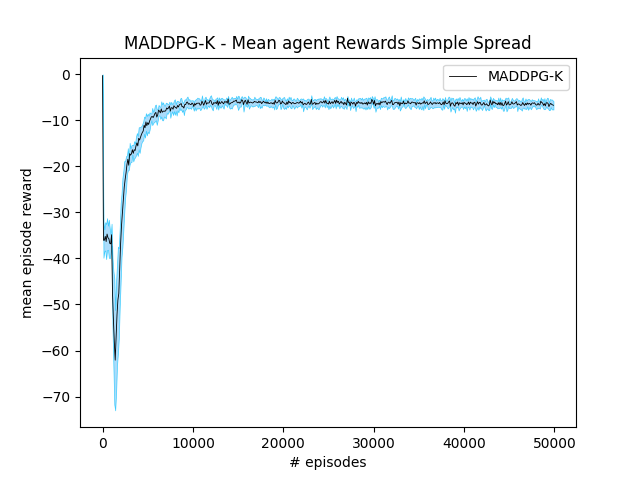}
    \caption{MADDPG and MADDPG-K Simple Spread N=3, K=2}
    \label{fig:maddpg_maddpg_k_simp_spread_n3}
\end{figure}
\begin{figure}[H]
    \centering
    \includegraphics[width=6cm]{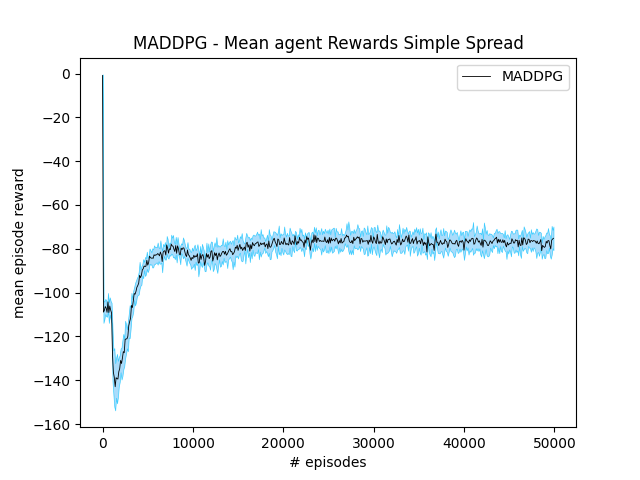}
    \includegraphics[width=6cm]{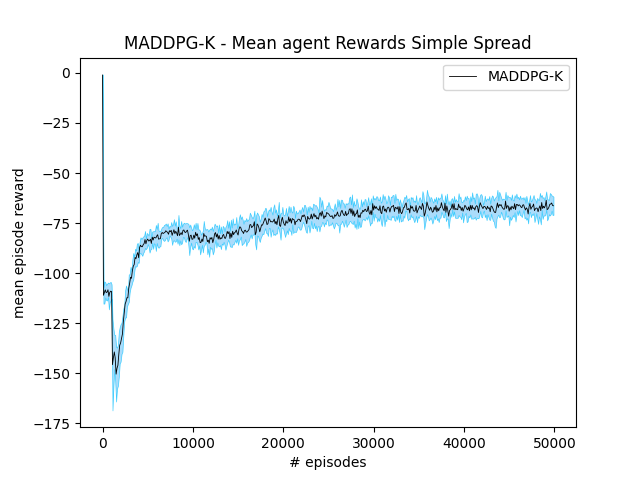}
    \caption{MADDPG and MADDPG-K Simple Spread N=5, K=2}
    \label{fig:maddpg_maddpg_k_simp_spread_n5}
\end{figure}\begin{figure}[H]
    \centering
    \includegraphics[width=6cm]{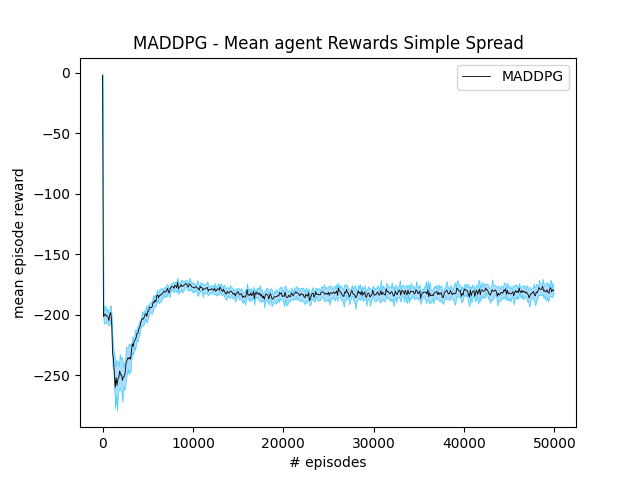}
    \includegraphics[width=6cm]{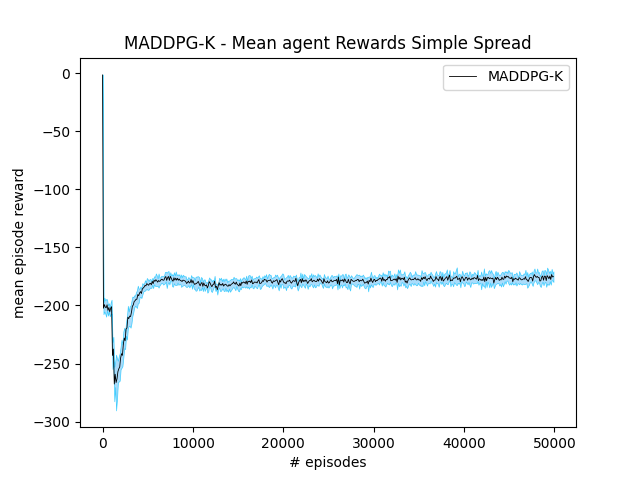}
    \caption{MADDPG and MADDPG-K Simple Spread N=7, K=2}
    \label{fig:maddpg_maddpg_k_simp_adv_n7}
\end{figure}\begin{figure}[H]
    \centering
    \includegraphics[width=6cm]{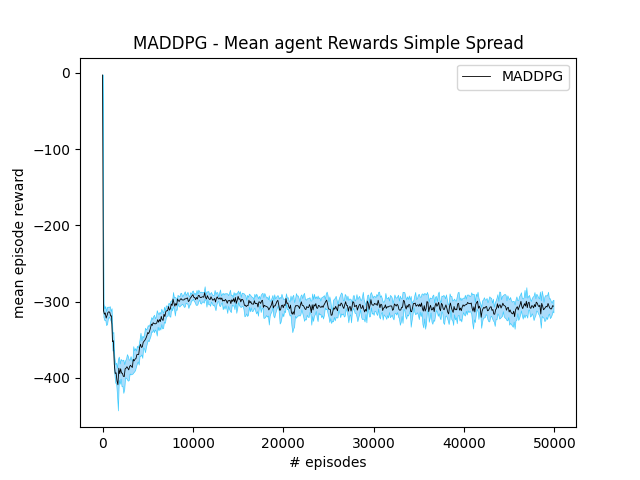}
    \includegraphics[width=6cm]{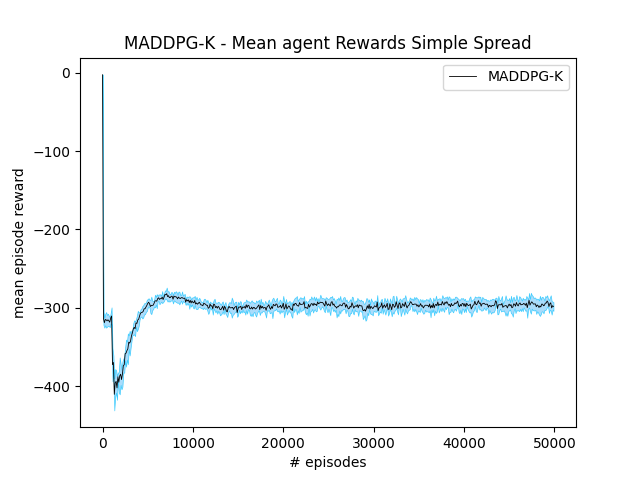}
    \caption{MADDPG and MADDPG-K Simple Spread N=9, K=2}
    \label{fig:maddpg_maddpg_k_simp_adv_n9}
\end{figure}

\begin{figure}
    \centering
    \textbf{Informal Practical Runtime Evaluation on our Hardware}
    \includegraphics[width=\textwidth]{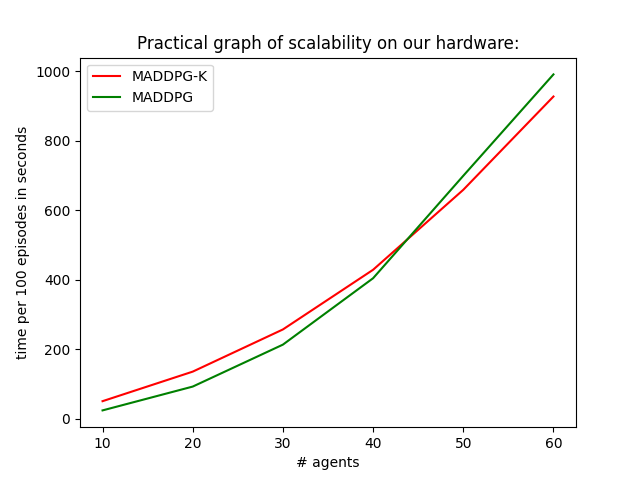}
    \caption{Scalability of MADDPG-K and MADDPG on Simple Spread environment}
    \label{fig:scalability}
\end{figure}\newpage
Note that in general because MADDPG-K by design focuses on a smaller neighborhood of agents, it will not require networks as large as MADDPG to learn an adequate policy. This additional factor is another important reason why MADDPG-K can help scale to environments with more agents. In the graph above we purposely use the same architecture for MADDPG and MADDPG-K to show that even with such a direct comparison MADDPG-K scales better.
\begin{figure}
    \centering
    \includegraphics{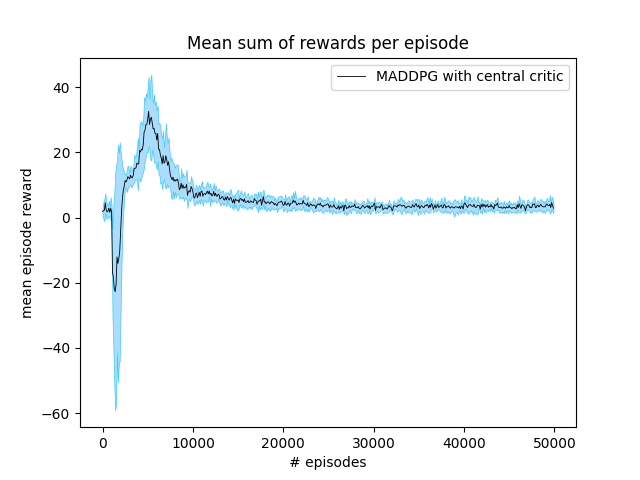}
    \caption{MADDPG with Central Critic Algorithm on Simple Tag}
    \label{fig:central_simp_tag}
\end{figure}\newpage
The convergence of the mean total reward across all agents to a value less than $10$ (around $5$ total average reward at convergence) clearly indicated that shared centralized critic approach as envisioned by us would not be viable. We can see that the total average reward is noticeably less than even the independent DDPG implementation achieved from the original MADDPG paper (around $10$ total average reward at convergence).

\subsection{Algorithms}
\begin{figure}
    \centering
    \textbf{The Original MADDPG Algorithm}
    \includegraphics[width=\textwidth]{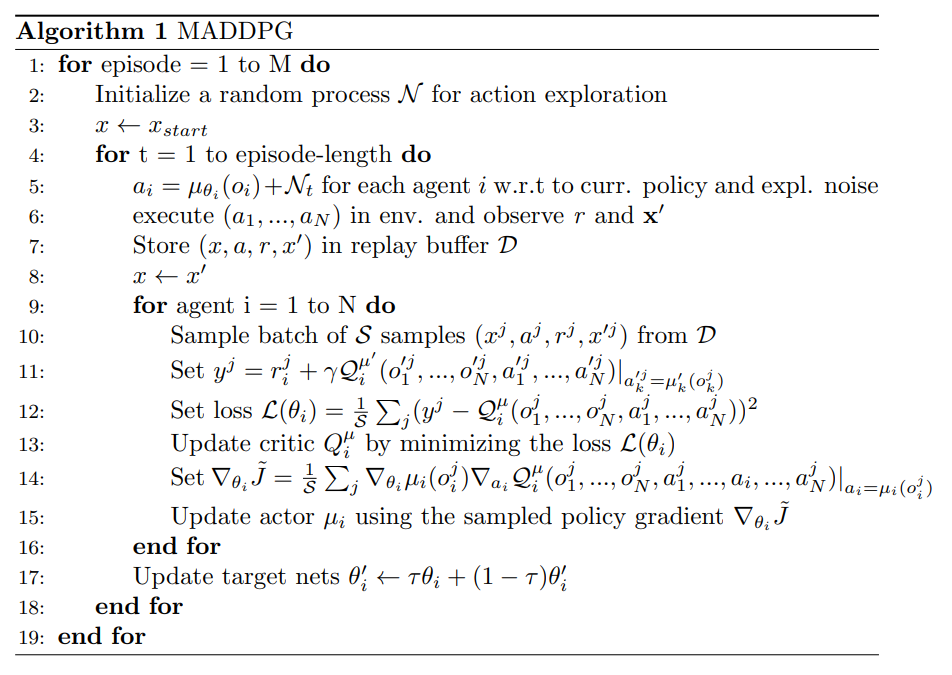}\\
    \caption{Original MADDPG algorithm}
    \label{fig:maddpg_orig_alg}
\end{figure}
Note that in the algorithm \ref{fig:maddpg_orig_alg}, $x = (o_1,...,o_N)$ is the vector of all observations and $x' = (o'_1,...,o'_N)$ the vector of all next observations.\\\\
Also note that $K$ can be defined for each agent type in our code however we have simplified this to one $K$ describing all the neighbors in our paper for the sake of simplicity.\newpage
\subsection{Methodology for Experiments}
\begin{itemize}
    \item Algorithms are tested by running them 10 times and then calculating the average reward trajectory throughout the learning process before plotting the obtained results. This methodology is followed for all the plots in this paper.
    \item In our paper $K$ is defined as being the neighbors excluding the agent itself however in our code $K$ is defined as being the neighbors including the current agent. Essentially in our paper $K=2$ is equivalent to $K=3$ in our code.
\end{itemize}
\end{document}